\documentclass[journal,twoside]{IEEEtran}

\usepackage[caption=false]{subfig}
\usepackage{cite}
\usepackage{url}
\usepackage{graphicx}
\usepackage[tbtags]{amsmath}
\usepackage{graphics}
\usepackage{epsfig}
\usepackage{amsthm, amssymb}
\usepackage{mathrsfs}
\usepackage{algorithm}
\usepackage{algorithmicx}
\usepackage{algpseudocode}
\usepackage[table,xcdraw]{xcolor}
\usepackage{array}
\usepackage{tabularx}

\newtheorem{theorem}{Theorem}

\usepackage{xspace}
\usepackage{marvosym}
\newcommand*{\QEDA}{\hfill\ensuremath{\blacksquare}}

\ifCLASSINFOpdf
\else
\fi

\begin{document}
	\title{Adaptive Split Learning over Energy-Constrained Wireless Edge Networks}
\author{\IEEEauthorblockN{
Zuguang~Li\IEEEauthorrefmark{1}\IEEEauthorrefmark{2},
Wen~Wu\IEEEauthorrefmark{1}$^{\textrm{,\;\Letter}}$,
Shaohua~Wu\IEEEauthorrefmark{2}\IEEEauthorrefmark{3},
and Wei~Wang\IEEEauthorrefmark{4}
}\\
\IEEEauthorblockA{\IEEEauthorrefmark{1}Frontier Research Center, Peng Cheng Laboratory, China} \\
\IEEEauthorblockA{\IEEEauthorrefmark{2}School of Electronics and Information Engineering, Harbin Institute of Technology, Shenzhen, China}\\
\IEEEauthorblockA{\IEEEauthorrefmark{3}Department of Broadband Communication, Peng
Cheng Laboratory, China} \\
\IEEEauthorblockA{\IEEEauthorrefmark{4}College of Electronic and Information Engineering, Nanjing University of Aeronautics and Astronautics, China}\\
Email: \{lizg01, wuw02\}@pcl.ac.cn, hitwush@hit.edu.cn, wei\_wang@nuaa.edu.cn
}

%
\maketitle

\begingroup\renewcommand\thefootnote{${\textrm{\Letter}}$}
\footnotetext{Wen Wu (wuw02@pcl.ac.cn) is the corresponding author of this paper. This paper has been published in \textit{IEEE INFOCOM 2024 - IEEE Conference on Computer Communications Workshops}.}
\endgroup

\IEEEpubidadjcol

\thispagestyle{empty}
\begin{abstract}
 Split learning (SL) is a promising approach for training artificial intelligence (AI) models, in which devices collaborate with a server to train an AI model in a distributed manner, based on a same fixed split point. However, due to the device heterogeneity and variation of channel conditions, this way is not optimal in training delay and energy consumption. In this paper, we design an adaptive split learning (ASL) scheme which can dynamically select split points for devices and allocate computing resource for the server in wireless edge networks. We formulate an optimization problem to minimize the average training latency subject to long-term energy consumption constraint. The difficulties in solving this problem are the lack of future information and mixed integer programming (MIP). To solve it, we propose an online algorithm leveraging the Lyapunov theory, named OPEN, which decomposes it into a new MIP problem only with the current information. Then, a two-layer optimization method is proposed to solve the MIP problem. Extensive simulation results demonstrate that the ASL scheme can reduce the average training delay and energy consumption by 53.7\% and 22.1\%, respectively, as compared to the existing SL schemes.

\end{abstract}

\section{Introduction}
The coming sixth generation (6G) network is expected to evolve from connecting people and things to connecting intelligence~\cite{shen2021Holistic}. At the same time, as the burgeoning edge devices and wireless sensing technology generate big data, edge computing will become an engine for edge network intelligence~\cite{deng2020edge}. The edge network intelligence deeply integrating wireless communications and artificial intelligence (AI) enables the edge devices for training AI models~\cite{xu2023edge,shen2020ai}. Privacy concerns and limited communication resources hinder edge devices from offloading their data to a central server for training AI models~\cite{xie2023mobfl}. 
Split learning (SL) is a promising approach to edge network intelligence, in which each participant trains only on its segment using its local data and transmits only the small-scale intermediate results at the split point rather than raw data. Therefore, SL can mitigate the aforementioned privacy and communication concerns to a significant extent~\cite{wu2023split}.

In recent years, SL has triggered a fast-growing research interest.
In \cite{kang2017neurosurgeon}, the per-layer execution time and energy consumption in different deep neural networks (DNNs) were analyzed, and then the optimal point could be selected for the best latency or best mobile energy.
In \cite{thapa2022splitfed}, an architecture called splitfed learning was proposed to address the issues in federated learning (FL) and SL, where all clients train their local device-side model in parallel and the fed server conducts the aggregation of the client-side local models.
In \cite{wang2022hivemind} a multi-split machine learning framework was proposed which reimagines the multi-split problem as a min-cost graph search task and optimally selects split points across multiple computing nodes to minimize the cumulative system latency.
Similarly, in~\cite{xu2023accelerating}, a split FL framework was designed to enhance communication efficiency for splitfed learning, where both the selection of split points and bandwidth allocation are jointly optimized to minimize the overall system latency.

However, in practical systems, due to the device heterogeneity and variation of channel conditions, static split points are adopted for the devices participating in model training, which is not optimal. In addition, the energy consumption performance should be considered in a system with constrained energy. 
Thus, in wireless edge networks, it is necessary to determine optimal split points for devices and computing resource allocation for the server, to minimize the training delay while satisfying long-term energy consumption constraint. 
The difficulties of this problem are multiple-fold. \textit{Firstly,} split point selection should dynamically adapt to the heterogeneous computing capabilities and uncertain channels during each training iteration. \textit{Secondly,} guaranteeing long-term energy consumption constraint requires future network information, e.g., channel conditions of devices and energy overhead of the system. Such information is unknown beforehand.

In this paper, \textit{firstly}, we design an adaptive split learning~(ASL) scheme aiming to address the device heterogeneity and channel uncertainty, and thus to reduce the model training delay. In ASL, the heterogeneous devices collaborate with an edge server to train an AI model. \textit{Secondly},
we formulate a joint optimization problem of the split point selection and computing resource allocation to minimize the average system latency subject to a long-term energy consumption constraint. \textit{Thirdly}, considering that future network information is required to solve this problem, we propose an \underline{O}nline s\underline{P}lit point selection and computing r\underline{E}source allocatio\underline{N} (OPEN) algorithm, which jointly determines the split point selection and computing resource allocation in an online manner only with current information. The OPEN algorithm based on the Lyapunov theory transforms the problem into a new problem only with the current information, and a two-layer optimization method is proposed to solve it.
\textit{Finally}, extensive simulations evaluate the performance of the proposed scheme. The main contributions of this work are summarized as follows: 
\begin{itemize}
	\item We propose an ASL scheme to reduce the model training delay in energy-constrained wireless edge networks, which addresses the device heterogeneity and channel uncertainty.
 
	\item We design an online algorithm (i.e., OPEN algorithm) to jointly determine the optimal split point and computing resource allocation decisions, and keep the average energy consumption within a threshold.
\end{itemize}

The remainder of this paper is organized as follows. Section~\ref{sec: modeling and Problem Formulation} introduces the system model and formulates the problem.
We elaborate on the OPEN algorithm in Section~\ref{sec: Proposed solution}. Section~\ref{sec: simulation results} presents the simulation results. Finally, Section~\ref{sec: conclusion} concludes this paper. 

\section{System Model and Problem Formulation} \label{sec: modeling and Problem Formulation}

\subsection{Considered Scenario}

\begin{figure}[t]
	\renewcommand{\figurename}{Fig.}
	\centering
	\includegraphics[width=0.4\textwidth]{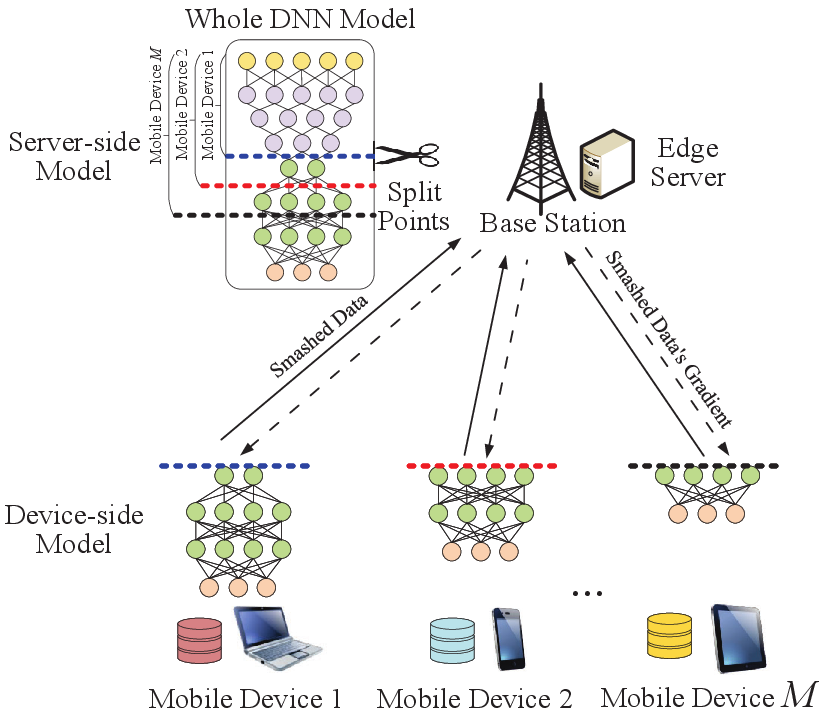}
	\caption{{System model.} }
	\label{fig:system_model}
    \vspace{-0.4cm}
\end{figure}

In this paper, a typical SL scenario is considered, in which multiple mobile devices (MDs) collaborate with a base station (BS) to train an AI model. Each participant only trains a portion of the whole AI model, as depicted in Fig.~\ref{fig:system_model}.
To complete model training process, the execution \emph{procedure} between the BS and MDs is as follows. 

First, the BS transmits the device-side model to a chosen MD, which then trains its own device-side model using its local data until reaching the split point, thereby producing smashed data. Second, this MD transmits the smashed data over the wireless channel to the BS, where it serves as input for the forward propagation on the remaining model layers. Third, at the BS, the edge server computes the loss by comparing the actual labels with the model's predictions and subsequently conducts backward propagation from the last layer up to the split point, yielding the gradient of the smashed data. Fourth, the BS relays the gradient of the smashed data at the split point back to the MD, enabling the MD to perform the backward propagation for its device-side model based on this gradient. Finally, upon completion of the local update, the MD transmits the updated device-side model to the BS, whereupon the device-side and server-side models are merged into an updated global model.
The steps above are repeated for all the participating MDs. A training episode refers to all the devices that have been trained one time. Multiple training episodes are performed until a satisfactory model performance is achieved.

The MDs are trained in a line manner, which are indexed by set $\mathcal{M}=\{1, 2, ..., M\}$. The model training episode is indexed by $n \in \mathcal{N}=\{1,2,..., N\}$. 
For the $m$-th MD participating in the SL, the split point decision at episode $n$ is denoted by $s_{m,n} \in \mathcal{S} =\{1,2,..., S\}, ~\forall m\in \mathcal{M}, ~n \in \mathcal{N}$,
where $\mathcal{S}$ represents the available split points in the considered AI model. This means that the front $s_{m,n}$ layers of the AI model are trained on the $m$-th MD, while the remained layers are trained on the edge server.
To guarantee the computing resource constraint, the edge computing resource allocated to the $m$-th MD at episode $n$ is denoted by $c_{m,n}$, satisfying
$0\leq c_{m,n}\leq 1,  ~\forall m\in \mathcal{M}, ~n \in \mathcal{N}$.

\subsection{Training Delay}
Executing SL requires model computation at the MD and BS and data exchange between them, which incurs computation delay and transmission delay. Both delay components are analyzed as follows.

\textit{{1) Computation delay:}} In each local update, the device-side model and the server-side model are trained at the MD and BS, respectively, whose delays are analyzed as follows:

\textit{Device-side model computation delay:} Let $\eta_D\left(s_{m,n}\right)$ denote the total computation workload of the $m$-th MD's device-side model at training episode $n$, where the amount of computation workload can be represented as the number of floating point operations (FLOPs) \cite{xu2019reform, xu2023accelerating}. 
Therefore, the computation delay of the device-side model can be defined as:
\begin{equation}\label{equ:Device-side model computation delay}
		d^{D,C}_{m,n}=\frac{\eta_D\left(s_{m,n}\right)}{F_m^D\delta_m^D\sigma_m^D},   ~\forall m\in \mathcal{M}, ~n \in \mathcal{N}.
	\end{equation}
Here, $F_m^D$ represents the processing unit frequency of the $m$-th MD, $\delta_m^D$ is the number of FLOPs processed by the $m$-th MD in a processing unit cycle depending on the architecture of the processor~\cite{zeng2021energy}, and $\sigma_m^D$ is the number of cores the $m$-th MD has.
	
\textit{Server-side model computation delay:} Let $\eta$ denote the total computation workload of the model. Consequently, the server-side model's computation workload is the difference, i.e., $\eta-\eta_D\left(s_{m,n}\right)$. Similar to that in \eqref{equ:Device-side model computation delay}, the computation delay of the server-side model is given by 
	\begin{equation}
		d_{m,n}^{B,C}=\frac{\eta-\eta_D\left(s_{m,n}\right)}{c_{m,n} F^B\delta^B \sigma^B},  ~\forall m\in \mathcal{M}, ~n \in \mathcal{N},
	\end{equation}
where $F^B$, $\delta^B$, and $\sigma^B$ represent the frequency of the computing server at the BS, the number of FLOPs processed by a cycle, and the number of cores, respectively. Here, $c_{m,n} F^B$ is the amount of allocated edge computing resources for the $m$-th MD.

\textit{{2) Transmission delay:}}
In SL, necessary data, MDs and the BS exchange essential data (e.g., device-side models, smashed data, and gradients) which incur transmission delays. The following provides a detailed analysis.

\textit{Device-side model downloading delay:} To cooperatively train the AI model with each MD, BS needs to send the latest device-side model to the MDs. Let
$R^{B}_{m,n} = W^B\log_2(1+ ({P^B \left| g^{B}_{m,n} \right|^2})/({(N_0+I)W^B}))$
represent the transmission rate from the BS to the $m$-th MD at episode $n$. Here, $W^B$, $P^{B}$, and  $g^{B}_{m,n}$ represent the downlink bandwidth, the transmission power at the BS, and the channel coefficient between BS and the $m$-th participating MD at episode $n$, respectively. $N_o$ and $I$ denote the power spectrum density of thermal noise and interference caused by other BSs, respectively. Denote the data size of the device-side model with split point $s_{m,n}$ as $\xi\left(s_{m,n}\right)$, which can be quantified by the count of parameters within that portion of the model.
Hence, we have the device-side model downloading delay as follows:
	\begin{equation}
		d_{m,n}^{B,D}=\frac{\xi\left(s_{m,n}\right)}{R^{B}_{m,n}},  ~\forall m\in \mathcal{M}, ~n \in \mathcal{N}.c	\end{equation}

\textit{Smashed data transmission delay:} In each local update, MDs send the smashed data to the BS for the forward propagation of the server-side model. The transmission rate from the $m$-th participating MD to the BS is represented by 
$R^{D}_{m,n}=W^D_{m}\log_2(1+({P^D_{m} \left| g^{D}_{m,n} \right|^2})/({(N_0+N)W^D_{m}}))$,
where $W^D_{m}$, $ g^{D}_{m,n} $, and $P^D_{m}$ represent the bandwidth, the channel coefficient from the $m$-th MD to the BS at episode $n$, and MD's transmission power in the uplink transmission, respectively. In addition, the size of the smashed data at the split point $s_{m,n}$ is $\beta(s_{m,n})$, which is measured by the number of parameters.
Thus, the smashed data transmission delay is given by
\begin{equation}
    	d_{m,n}^{D,S}=\frac{\beta(s_{m,n})}{R^{D}_{m,n}},  ~\forall m\in \mathcal{M}, ~n \in \mathcal{N}.
\end{equation}

\textit{Gradient transmission delay:} The BS needs to send the smashed data's gradient to MDs for the device-side model's backward propagation. Let $\gamma(s_{m,n})$ denote the data size of smashed data's gradient given the split point $s_{m,n}$, which is also measured by the number of parameters.
The corresponding delay is calculated as follows:
\begin{equation}
	d_{m,n}^{B,G}=\frac{\gamma(s_{m,n})}{R^{B}_{m,n}},  ~\forall m\in \mathcal{M}, ~n \in \mathcal{N}.
\end{equation}

\textit{Device-side model uploading delay:} After a few local updates, the MD needs to upload the latest device-side model to the BS for training the AI model with the next MD. The corresponding delay is given by
\begin{equation}
        d_{m,n}^{D,D}=\frac{\xi\left(s_{m,n}\right)}{R^{D}_{m,n}},  ~\forall m\in \mathcal{M}, ~n \in \mathcal{N}.
\end{equation}

\textit{{ 3) Overall model training delay:}}
Taking all the computation and communication delay components into account, the model training delay taken for the $m$-th participating MD at each episode is given by
\begin{equation}\label{equ:model training delay}
	\begin{split}
		D\left(s_{m,n}, c_{m,n}\right)& = d^{D,C}_{m,n} + d^{B,C}_{m,n} + d^{B,D}_{m,n} + d^{D,S}_{m,n} \\
		&   + d^{B,G}_{m,n} + d^{D,D}_{m,n},  ~\forall m\in \mathcal{M}, ~n \in \mathcal{N}.
	\end{split}
\end{equation}

The performance of a model is intricately linked to the batch size and number of episodes used during training.  A model requires approximately 100 episodes to achieve satisfactory performance, especially when there are a large amount of participating MDs. The overall model training times, denoted by $MN$, can be considered as long terms. The average model training delay across all MDs and the BS until a satisfactory model performance is achieved, defined as $\overline{D}$, can be represented by
\begin{equation} \label{average_delay}
	\overline{D} = \lim_{MN \to \infty}\frac{1}{MN}\sum_{n=1}^{N} \sum_{m=1}^{M} D\left(s_{m,n}, c_{m,n}\right).
\end{equation}

\subsection{Energy Consumption}
In SL, all participating MDs need to work with the edge server to complete the model training, which imposes a heavy energy consumption burden on MDs, the BS, and its edge server. The split point selections and computation resource allocation decisions will incur different energy consumption in this system. 
The energy consumption of the system consists of the components as follows.

\textit{{ 1) Transmission-related energy consumption:}} It is caused by data transmission, including device-side model, smashed data, and gradients. The energy consumption of the $m$-th MD is a product to the MD's transmission power and communication period, i.e., 
    \begin{equation}
		E^{D,T}_{m,n} = P_{m}^D \left(d_{m,n}^{D,S}+ d_{m,n}^{D,D}\right).
	\end{equation}
 Similarly, the energy consumption of the BS is denoted as
	\begin{equation}
		E^{B,T}_{m,n} = P^B\left(d_{m,n}^{B,D}+ d_{m,n}^{B,G}\right).
	\end{equation}
 
\textit{{ 2) Computation-related energy consumption:}} It is caused by the model training. The energy consumption for each central processing unit (CPU) cycle can be represented by $ \kappa F^2$, where $\kappa$ reflects the effective switched capacitance, a parameter that depends on the chip architecture~\cite{chen2020joint, dai2018joint}.
The energy consumption for device-side and sever-side model training can be defined as
    \begin{equation}
   	\begin{split}
   		E^{D,C}_{m,n} &= \kappa \delta_m^D \sigma_m^D \left( F_m^D\right)^2 \eta_D\left(s_{m,n}\right),
   	\end{split}
   \end{equation}
   and
   \begin{equation}
   	\begin{split}
   		E^{B,C}_{m,n} &= \kappa c_{m,n} \delta^B \sigma^B \left( F^B\right)^2 \left(\eta-\eta_D\left(s_{m,n}\right) \right).
   	\end{split}
   \end{equation}
   
\textit{{ 3) Overall model energy consumption:}}
All energy consumption of the $m$-th MD at episode $n$ is $E^D_{m,n}$, denoted as $E^D_{m,n} =  E^{D,T}_{m,n} + E^{D,C}_{m,n}$. For the BS, the overall energy consumption at episode $n$ is $E^B_{m,n}$, denoted as $E^B_{m,n} = E^{B,T}_{m,n} + E^{B,C}_{m,n}$.
Taking all energy consumption of the $m$-th MD and BS into account, the overall energy consumption at each episode is given by
\begin{equation}\label{equ:overall_energy_consumption}
    \begin{split}
	E(s_{m,n}, c_{m,n}) & = E^D_{m,n} + E^B_{m,n},  ~\forall m\in \mathcal{M}, ~n \in \mathcal{N}.
    \end{split}
\end{equation}
Similar to (\ref{average_delay}), the average energy consumption of the MDs and BS until achieving a satisfactory model performance, defined as $\overline{D}$, can be represented by 
\begin{equation} \label{average_energy_consumption}
	\overline{E} = \lim_{MN \to \infty}\frac{1}{MN}\sum_{n=1}^{N} \sum_{m=1}^{M} E(s_{m,n}, c_{m,n}).
\end{equation}
In the following optimization problem, we keep the average energy consumption within a threshold to save the edge energy. 

\subsection{Problem Formulation}
We jointly determine split point selection and edge computing resource allocating decisions to minimize the average model training delay while satisfying the average energy consumption requirement, which is formulated as the following problem:
\begin{equation}\label{Problem 1}
    \begin{split}
    {\mathbf{P}_1:\;}	 \underset{\{s_{m,n}, c_{m,n}\} }{\text{min} }~& \overline{D}  \\
    \text{s.t.}~& \overline{E} \leq E_{th},  \\
    & ~ 0 \leq c_{m,n}\leq 1, ~\forall m\in \mathcal{M}, ~n \in \mathcal{N}, \\
    & ~ s_{m,n}\in \mathcal{S}, ~\forall m\in \mathcal{M}, ~n \in \mathcal{N}.
    \end{split}
\end{equation}
Here, the first constraint requires that the system's average energy consumption does not surpass upper limit $E_{th}$. The second and third constraints guarantee the feasibility of edge computing resource allocation and split point selection. 

Problem $\mathbf{P}_1$ is a long-term stochastic optimization problem due to the time-varying channel conditions. In addition, obtaining the optimal policy requires the complete future information, which is impossible. Even if the future information is available, the problem is a mixed integer programming (MIP) since the split point selection is discrete whereas the edge computing resource allocation is continuous.

\section{OPEN Algorithm} \label{sec: Proposed solution}
\subsection{Problem Transformation}
The primary obstacle in tackling problem $\mathbf{P}_1$ lies in handling the long-term constraints. To overcome this hurdle, we employ the Lyapunov optimization method \cite{neely2010stochastic, luo2019adaptive}. The essence of this method involves constructing energy deficit queues to represent the fulfillment status of long-term energy consumption constraints, thereby steering the system towards adhering to these constraints. The process of problem transformation is presented as follows. 

First, let $T$ denote the total number of model training times between the BS and MDs, i.e., $T = MN$, indexed by $t \in \mathcal{T}=\{1, 2, 3, ..., T\}$. To satisfy the energy consumption limitation, we introduce an energy deficit queue for the system with its dynamic evolves as follows:
\begin{equation} \label{eq: energy deficit queue}
	Q^{t+1}_{m,n}=\left[Q^{t}_{m,n}+E^{t}_{m,n} -E_{th}\right]^+, ~\forall m\in \mathcal{M}, ~n \in \mathcal{N}.
\end{equation}
Here, $Q^{t}_{m,n}$ indicates the queue backlog in time slot $t$ representing the accumulated part of current energy consumption that exceeds the upper limit, and $E^{t}_{m,n}$ is the corresponding overall energy consumption $E(s_{m,n}, c_{m,n})$ in the time slot $t$. The initial state is set to $Q^{0}_{m,n} = 0$ and the queue is updated according to \eqref{eq: energy deficit queue}.

\begin{theorem} Equation \eqref{eq: energy deficit queue} is essentially equivalent to the long-term throughput constraint in \eqref{Problem 1}, if the stability condition $ \lim_{T \to \infty}  Q^{T}_{m,n}/T = 0, ~ \forall m\in \mathcal{M}, ~n \in \mathcal{N}$ can be satisfied.
\label{theorem: 1}
\end{theorem}

\noindent \emph{Proof.}
	Please refer to the Proof of Theorem 1 in an online appendix \cite{appendix}.
\QEDA

Second, we utilize a Lyapunov function, defined as $L(Q^{t}_{m,n}) = {(Q^{t}_{m,n})^2}/2$,  to capture the fulfillment status of the long-term energy consumption constraint~\cite{xu2018joint}. The Lyapunov function serves as a quantitative indicator of congestion across all queues. A smaller value of $L(Q^{t}_{m,n})$ signifies a lesser queue backlog, thus indicating enhanced stability of the virtual queue.
$L(Q^{t}_{m,n})$ is large when queue backlogs are elevated, resulting in more flows experiencing throughput below the relevant required values. To ensure the energy deficit queue remains stable, i.e., to continuously enforce the energy consumption constraints by driving down the Lyapunov function, we introduce \textit{one-shot Lyapunov drift}, defined as
\begin{equation} \label{eq: one-shot Lyapunov drift}
	\bigtriangleup(Q^{t}_{m,n}) = \mathbb{E}[L(Q^{t+1}_{m,n}) - L(Q^{t}_{m,n}) | Q^{t}_{m,n}].
\end{equation} 
Then, we have 
 \begin{equation}\label{eq: Lyapunov drift}
    \begin{split}
	\bigtriangleup(Q^{t}_{m,n}) & = \frac{1}{2}\mathbb{E}\left[(Q^{t+1}_{m,n})^2 -(Q^{t}_{m,n})^2 | Q^{t}_{m,n}\right]  \\
    & \le \frac{1}{2}\mathbb{E}\left[\left(Q^{t}_{m,n}+E_{m,n}^{t} -E_{th}\right)^2 
    -(Q^{t}_{m,n})^2 | Q^{t}_{m,n}\right] \\
    & = B_1 
    + Q^{t}_{m,n} \mathbb{E}\left[\left(E_{m,n}^{t} -E_{th}\right) |  Q^{t}_{m,n}\right]  \\
    & \le B_2 + Q^{t}_{m,n} \mathbb{E}\left[\left(E_{m,n}^{t} -E_{th}\right) |  Q^{t}_{m,n}\right], 
    \end{split}
\end{equation}
where $B_1 = \left(E_{m,n}^{t} -E_{th}\right)^2/2$, and $B_2 = \left(E_{m,n}^{\max} -E_{th}\right)^2 / 2$ is a constant. $E_{m,n}^{\max}$ is the maximum energy that can be produced by the $m$-th MD and BS at episode $n$. The first inequality in \eqref{eq: Lyapunov drift}, is due to $(Q^{t}_{m,n})^2 \le \left(Q^{t}_{m,n}+ E_{m,n}^{t} -E_{th}\right)^2$, and the second inequality is because $E_{m,n}^{t} \le E_{m,n}^{\max}$.

Third, leveraging the Lyapunov optimization theory, the original problem $\mathbf{P}_1$ can be reformulated into minimizing a \textit{drift-plus-cost} at each time slot, as expressed below:
\begin{equation}\label{eq: drift-plus-cost}
\begin{split}
& \bigtriangleup(Q^{t}_{m,n}) + V \mathbb{E} \left[ D^{t}_{m,n} | Q^{t}_{m,n}\right]\\
& \le B_2 + Q^{t}_{m,n} \mathbb{E}\left[\left(E_{m,n}^{t} -E_{th}\right) |  Q^{t}_{m,n}\right] 
 + V \mathbb{E} \left[ D^{t}_{m,n} | Q^{t}_{m,n}\right],
\end{split}
\end{equation}
Here, $D^{t}_{m,n}$ represents the overall energy consumption $D(s_{m,n}, c_{m,n})$ at the time slot $t$, and $V$ is a positive and adjustable parameter that adjusts the balance between the training delay and energy consumption.  The inequality in \eqref{eq: drift-plus-cost} is due to the upper limit in  \eqref{eq: Lyapunov drift}.

Because both $B_2$ and $E_{th}$ are constants, the original problem $\mathbf{P}_1$ can be reformulated into a new optimization problem $\mathbf{P}_2$:
\begin{equation}	\label{Problem 2}
	\begin{split}
		{\mathbf{P}_2:\;}	 \underset{\{s_{m,n}^t, c_{m,n}^t\}_{\substack{\forall m\in \mathcal{M}, \\ ~n \in \mathcal{N}}} }{\text{min} }~~  &  f \\
		\text{s.t.}~&  0 \leq c_{m,n}^t \leq 1, ~\forall m\in \mathcal{M}, ~n \in \mathcal{N}, \\
		& s_{m,n}^t\in \mathcal{S}, ~\forall m\in \mathcal{M}, ~n \in \mathcal{N},
	\end{split}
\end{equation} 
where $f = V D_{m,n}^{t} +Q^{t}_{m,n} E_{m,n}^{t}$.
It is worth noting that solving problem $\mathbf{P}_2$ only requires the current information as input.  

Problem $\mathbf{P}_2$ is a non-convex optimization problem, in which two decision variables are mixed integers and the objective function is non-convex. To solve the MIP problem, we propose a two-layer optimization method to jointly determine the optimal split point and computing resource allocation decisions, which is detailed as follows.

\subsection{Upper Layer Analysis}
When the split point selecting decision is fixed, the edge computing resource allocation problem can be transformed into an upper-layer problem as follows:
\begin{equation}	\label{Problem 3}
	\begin{split}
		{\mathbf{P}_3:\;}	 {\text{min} }~~  &  f\\
		\text{s.t.}~&  0 \leq c_{m,n}^t\leq 1.
	\end{split}
\end{equation} 

\begin{theorem} For a given split point selecting decision $\{s_{m,n}^t\}$, the optimal solution for problem $\mathbf{P}_3$ is given by
\begin{equation}
	(c_{m,n}^t)^\star =
	 \begin{cases}
		1, \hfill   \text{if } \sqrt{\frac{V\omega_1}{\omega_4 Q^{t}_{m,n}}}>1, \\
		\sqrt{\frac{V\omega_1}{\omega_4 Q^{t}_{m,n}}}, \hfill \text{otherwise},
	\end{cases}
 \label{eq: optimization c_mi}
\end{equation}
where $\omega_1={\left(\eta-\eta_D\left(s_{m,i}^t\right)\right)}/{\left(F^B\delta^B \sigma^B\right)}$, and $\omega_4=\kappa \delta^B \sigma^B \left( F^B\right)^2 \left(\eta-\eta_D\left(s_{m,i}^t\right)\right)$.

\label{theorem: 2}
\end{theorem}

\noindent \emph{Proof.}
Please refer to the Proof of Theorem 2 in the online appendix.~\QEDA

\subsection{Lower layer Analysis}
When the optimal computing resource allocation is given, the split point selection problem can be transformed into a lower-layer problem as follows:
\begin{equation}	\label{Problem 4}
	\begin{split}
		{\mathbf{P}_4:\;}	 {\text{min} }~~  &  f \\
		\text{s.t.}~&  s_{m,n}^t \in \mathcal{S}.
	\end{split}
\end{equation}
According to ~\eqref{equ:model training delay} and ~\eqref{equ:overall_energy_consumption},
the objective function in ~\eqref{Problem 4} can be rewritten as
 \begin{equation}\label{equ: objective function 2}
 	\begin{split}
 		 	f\left(s_{m,n}^t \right) &= V D_{m,n}^{t} + Q^{t}_{m,n} E_{m,n}^{t}\\
 		 	&= V   \left( d^{D,C}_{m,n} + d^{B,C}_{m,n} + d^{B,D}_{m,n} + d^{D,S}_{m,n}
		              + d^{B,G}_{m,n} + d^{D,D}_{m,n} \right) \\
            & ~~  + Q^{t}_{m,n} \left(E^{B,T}(s_{m,n}^t) + E^{B,C}(s_{m,n}^t, c_{m,n}^t)\right).
 	\end{split}
 \end{equation}
The objective function in problem $\mathbf{P}_4$ is non-convex because the data sizes (e.g., device-side model, smashed data, and its gradient) are arbitrary functions in terms of the split point. This renders an analytical closed-form expression for the optimal split point infeasible, necessitating numerical methods for its calculation. Given the finite number of potential split points within the AI model, the lower-layer problem can be tackled using an exhaustive searching method.

The OPEN algorithm is shown in Algorithm~\ref{OPEN algorithm}. In the algorithm, the local optimal decision of the edge computing resource allocation is determined based on \eqref{eq: optimization c_mi}, and the local optimal split point is obtained by searching those split points to minimize $f\left(s_{m,n}^t \right)$. Then, the optimal split point and computing resource allocation decisions can be determined through several iterations. 

\begin{algorithm}[t]\small
	\caption{{OPEN algorithm}}
	\label{OPEN algorithm}
	\begin{algorithmic}[1]
		\Require
        $Q^{0}_{m,n} \gets 0, F_m^D, \delta_m^D, \sigma_m^D, F^B, \delta^B, \sigma^B, P_m^D,$
        $P^B, E_{th}$;
		\Ensure
		split point selecting decisions $(s_{m,n}^t)^{\star}$, and computing resource allocating decisions $(c_{m,n}^t)^{\star}$;
		\State Initiate $R_{m,n}^{D,t}$,  $R_{m,n}^{B,t}$, $f_{min}$, $Q^{t-1}_{m,n}$;
        \State Set $(c_{m,n}^t)^{\star} = 1$, and $ (s_{m,n}^t)^{\star} = 0$;
        \While{ $|| (c_{m,n}^t)^{\star} - c_{m,n}^{t, last}|| > 0.01$ and $(s_{m,n}^t)^{\star} \neq s_{m,n}^{t,last}$}
        \State $c_{m,n}^{t, last} = (c_{m,n}^t)^{\star}, s_{m,n}^{t,last} = (s_{m,n}^t)^{\star}$;
        \State Compute $(c_{m,n}^t)^{\star}$ based on \eqref{eq: optimization c_mi};
        \For{$ s_{m,n}^t = 1$ to $S$}
        \State Compute $D_{m,n}^t$ based on \eqref{equ:model training delay};
        \State Compute $E_{m,n}^t$ based on \eqref{equ:overall_energy_consumption};
        \State $Q^{t}_{m,n}=\max\{ Q^{t-1}_{m,n}+E^{t}_{m,n} -E_{th} ,0 \}$;
        \State Compute $f\left(s_{m,n}^t \right)$ based on \eqref{equ: objective function 2};
        \If{$f\left(s_{m,n}^t\right) < f_{min}$}
        \State  $f_{min} = f\left(s_{m,n}^t\right)$;
        \State $(s_{m,n}^t)^{\star} = s_{m,n}^t$;
        \EndIf
        \EndFor
        \EndWhile
        \State \Return $(s_{m,n}^t)^{\star}, (c_{m,n}^t)^{\star}$
	\end{algorithmic}
\end{algorithm}

\section{Simulation Results}\label{sec: simulation results}
\subsection{Simulation Setup}

The time-varying wireless channel gain of the $m$-th MD, i.e., $g^{D}_{m,n}$, is derived from a Rayleigh fading model as $ g^{D}_{m,n} = \rho^{D}_{m,n}  \Bar{g}^{D}_{m} $ \cite{liang2019deep}. Here, $\rho^{D}_{m,n}$ is an independent random channel fading factor following an exponential distribution with unity mean. Meanwhile, $\Bar{g}^{D}_{m} $ is the average channel gain calculated by using the free-space path loss model as $\Bar{g}^{D}_{m} = A_m^D \left(  \frac{3 \cdot 10^8}{4\pi f_m^D d_m} \right)^{\phi_m^D}$. Here, $A_m^D$, $f_m^D$, and $\phi_m^D$ are the antenna gain, the carrier frequency, and the path loss exponent of the $m$-th MD, respectively, while $d_m$ is the distance between the $m$-th MD and BS. Analogous to $g^{D}_{m,n}$, the BS's time-varying wireless channel gain $g^{B}_{m,n}$ also adheres to the same fading channel model, but it incorporates different values for some parameters.

In the simulation, we adopt a 12-layer LeNet model~\cite{nishio2019client} with the widely-adopted MNIST data. The batch size of the training model is set to be $16$.
Due to the small amount of computation workload and parameters in the activation layers (ReLU), we restrict the selection of split points to only include convolution layers and fully-connected layers. Thus, there are 12 available split points in the LeNet model.

We consider 30 MDs and one BS with an edge server, as illustrated in Fig.~\ref{fig:system_model}. Here, the MDs' processing unit frequencies follow a uniform distribution within $\left[0.5, 3\right]$ GHz, the numbers of cores the MDs have follow a uniform distribution within $\left[1, 8\right]$, and the distances between MDs and the BS follow a uniform distribution within $\left[100, 1000\right]$ meters. Specially, the processing unit frequencies of MD 1, MD 2, MD 3, and MD 4 are set to 0.5 GHz, 1 GHz, 2 GHz, and 4 GHz, their numbers of cores are 1, 4, 8, and 16, respectively, and their distances are all set to 200 meters. The other main simulation parameters are concluded in Table~\ref{Table: Simulation parameters}.

\begin{table}[t]
	\scriptsize
    \renewcommand\arraystretch{1.5}
	\centering
	\caption{Simulation parameters.}
	\label{Table: Simulation parameters}
			\vspace{-0.2cm}
	\begin{tabular}{cc|cc}
		\hline
        \hline
	  \textbf{Parameter} & \textbf{Value}  &  \textbf{Parameter} & \textbf{Value} \\
        \hline
        $A_m^D$  & 4.11 & $A_m^B$  & 8 \\
        $\delta_m^D$  & 8 & $\delta^B$  & 16  \\
        
         $\sigma^B$  & 32 & $\kappa$ & $10^{-26}$ \\
         
        $\phi_m^D$ & 1 & $\phi_m^B$ & 1 \\
        $F^B$  & 3 GHz & $E_{th}$ & 3,000 J \\
        $f_m^D$  & 2,000 MHz & $f_m^B$ & 2,000 MHz \\
        $P_m^D$  & 0.4 W & $P^B$  & 3 W \\
        $W_m^D$ & 20 MHz & $W^B$  & 40 MHz \\ 
        
	    $N_o$ & $-174$ dBm/Hz & $N$ & $-164$ dBm/Hz \\

        \hline
	\end{tabular}
	\vspace{-0.4cm}
\end{table}	

\subsection{Performance Evaluation}

\begin{figure}[t]
	\centering
	\subfloat[Optimal split point.]
    {
        \centering
    	\includegraphics[width=0.23\textwidth]{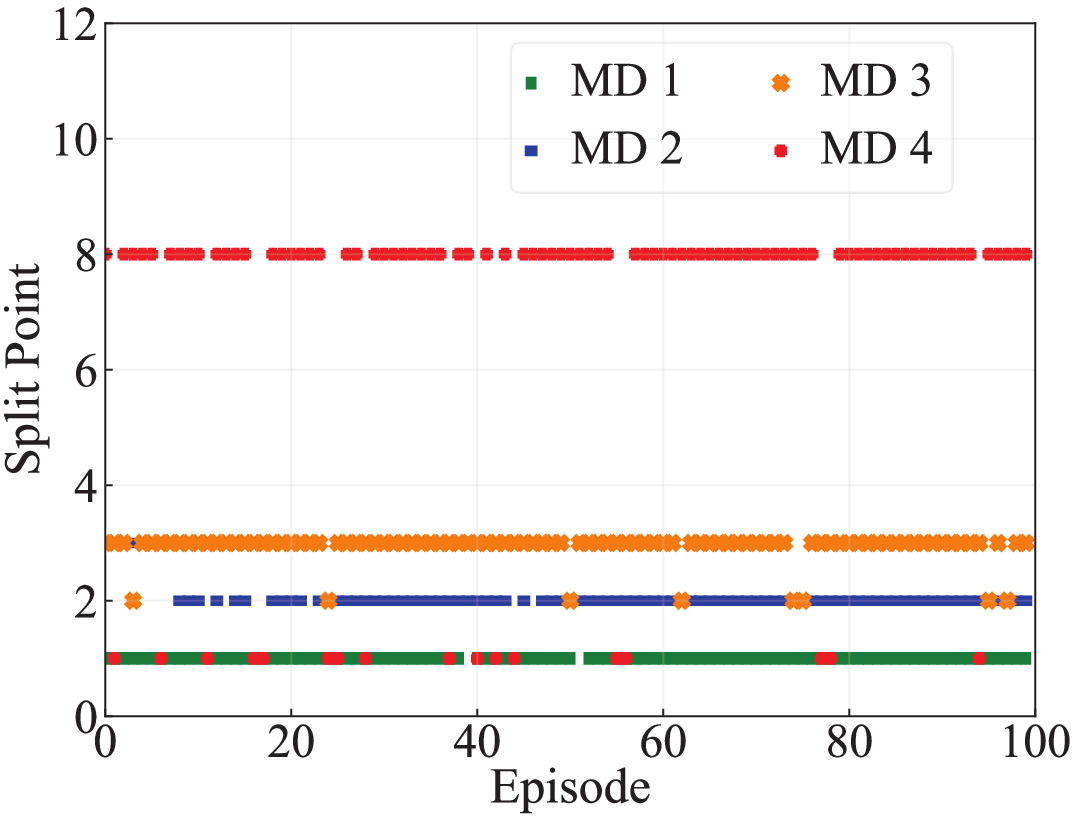}
    	\label{fig:sub1}
    }
	\subfloat[Optimal resource allocation decision.]
    {
        \centering
    	\includegraphics[width=0.23\textwidth]{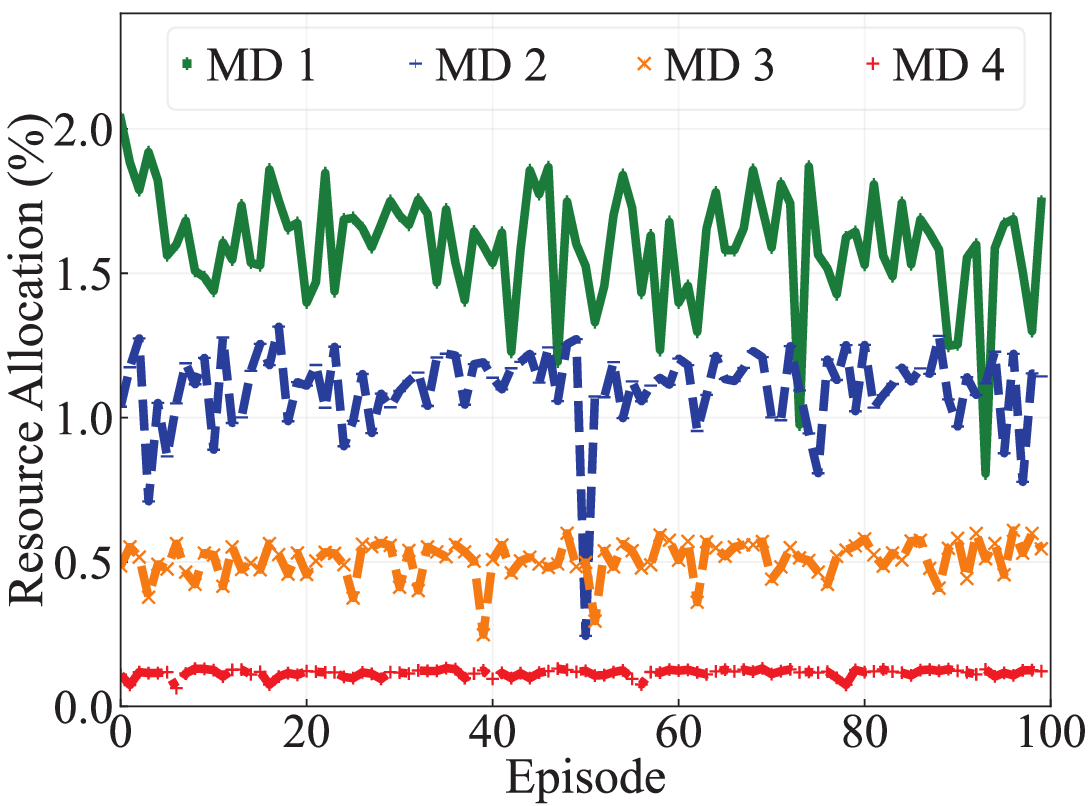}
        \label{fig:sub2}
    }
	\caption{ Optimal split point and computing resource allocation at each episode.}
	\label{fig: split and resource with episode}
 \vspace{-0.4cm}
\end{figure}

We evaluate the optimal split points and computing resource allocation decisions on the MDs with different computing power. 
As shown in Fig.~\ref{fig:sub1}, we can observe that the optimal split points of MDs dynamically change with episodes, but each of them maintains a split point in general. Furthermore, with the progressive enhancement of an MD's computing capability, its optimal split point shifts closer to the output end.  Consequently, it is a lesser need for allocating computing resources, as depicted in Fig.~\ref{fig:sub2}. This is because the MD with strong computing power prefers training the majority of the AI model locally rather than delegating it to the edge server, which benefits reducing the transmission delay. 

\begin{figure}[t]
    \vspace{-0.5cm}
	\centering
	\subfloat[Delay across all episodes.]
    {
        \centering
    	\includegraphics[width=0.23\textwidth]{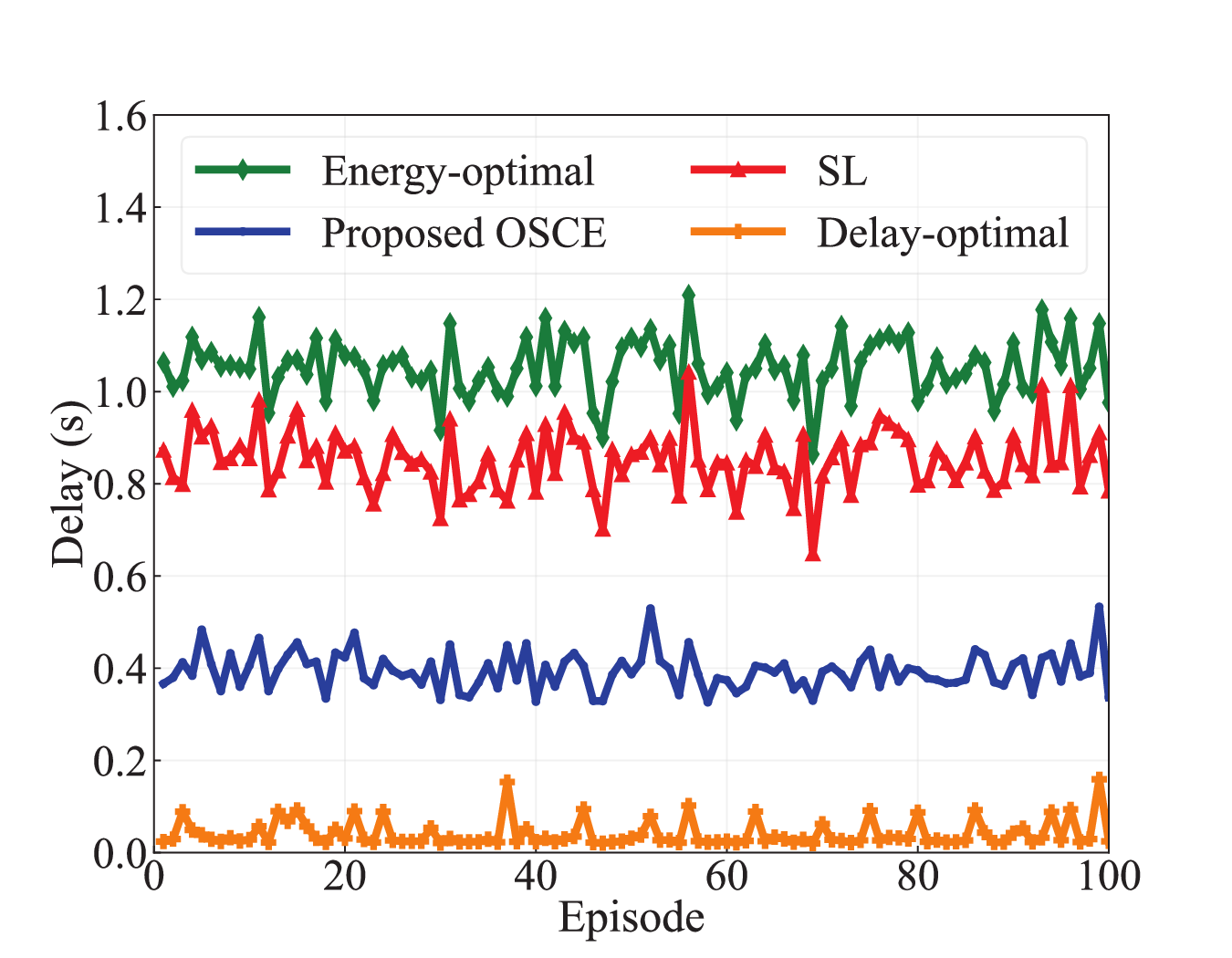}
    	\label{fig2:sub1}
    }
	\subfloat[Energy consumption across all episodes.]
    {
        \centering
    	\includegraphics[width=0.23\textwidth]{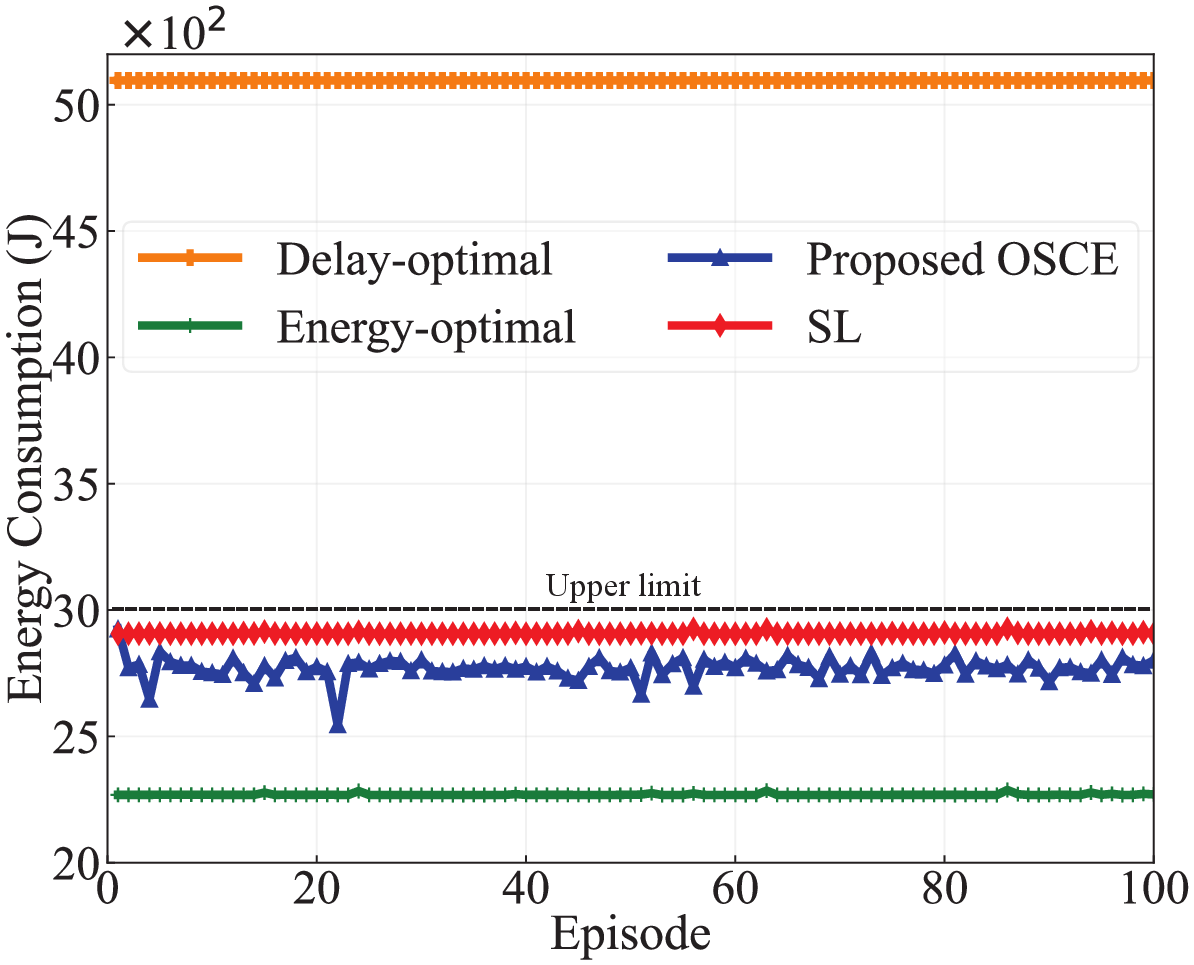}
        \label{fig2:sub2}
    }
	\caption{Delay and energy consumption performance with respect to different schemes.}
	\label{fig2: split and resource with episode}
 \vspace{-0.2cm}
\end{figure}

We compare the ASL scheme with the following benchmark schemes: (1) \textit{SL,} where all devices train the AI model before split point 9 and the edge server releases the total computing resource; (2) \textit{Delay-optimal,} where the goal is to minimize the overall model training delay; (3) \textit{Energy-optimal,} where minimizing the overall energy consumption is the goal.
As shown in Fig.~\ref{fig2:sub1}, the delays under different schemes fluctuate with the model training, but all remain stable in general. We can observe that the proposed ASL scheme achieves 53.7\% and 62.5\% reduction in the average training delay, as compared to the SL and energy-optimal schemes, respectively.
Although the training delay of delay-optimal scheme is smaller than that of the ASL scheme, its energy consumption is much larger than that of the ASL scheme, as shown in Fig.~\ref{fig2:sub2}. In addition, from Fig.~\ref{fig2:sub2}, we can see that the energy consumption of the ASL scheme is always maintained below the upper limit, i.e., 3,000~J. Compared to SL scheme, the average energy consumption of the proposed ASL scheme is reduced by 22.1\%.

\section{Conclusion}\label{sec: conclusion}
In this paper, we have designed a novel ASL scheme for deploying the AI model in energy-constrained wireless edge networks. We have formulated a problem to minimize long-term system latency subject to a long-term energy consumption constraint. To solve this problem, we have proposed an online algorithm to make the optimal split point and computing resource allocation decisions. Extensive
simulation results have demonstrated the effectiveness of the ASL scheme in reducing training latency. Due to low training latency and acceptable energy consumption, the ASL scheme can be applied to facilitate AI model training
in energy-constrained wireless networks. For future work, we will investigate the split point selection problem in non-linear AI models.

\bibliographystyle{IEEEtran}
\bibliography{main}

\end{document}